\documentclass[runningheads]{llncs}

\usepackage{eccv}

\usepackage[T1]{fontenc}
\usepackage[utf8]{inputenc}
\usepackage{graphicx}
\usepackage{xcolor}
\usepackage{amsmath,amssymb}
\usepackage{booktabs}
\usepackage{multirow}
\usepackage{array}
\usepackage{tabularx}
\usepackage{siunitx}
\usepackage{enumitem}
\usepackage{float}
\usepackage{subcaption}
\usepackage{caption}
\usepackage{url}
\usepackage{hyperref}
\usepackage{booktabs}

\hypersetup{
    colorlinks=true,
    linkcolor=blue,
    citecolor=blue,
    urlcolor=blue
}
\usepackage{eccvabbrv}

\usepackage[accsupp]{axessibility}  % Improves PDF readability for those with disabilities.

\usepackage{hyperref}

\usepackage{orcidlink}

\begin{document}

\title{Author Guidelines for ECCV Submission} 


% % TODO FINAL: Replace with your author list. 
% % Include the authors' OCRID for the camera-ready version, if at all possible.
% Second Author\inst{2,3}\orcidlink{1111-2222-3333-4444} \and
% Third Author\inst{3}\orcidlink{2222--3333-4444-5555}}

% % TODO FINAL: Replace with an abbreviated list of authors.

% % TODO FINAL: Replace with your institution list.
% Springer Heidelberg, Tiergartenstr.~17, 69121 Heidelberg, Germany

\title{AGT-CV: An Aerial-Ground Team Cross-View Dataset for Heterogeneous Robot Teams in Unstructured Environments}
\titlerunning{AGT-CV: An Aerial-Ground Team Cross-View Dataset} % shorter version for the page header, otherwise we get "title suppressed due to excessive length" on every other page

\author{
Siwei Cai\textsuperscript{*}%\inst{1}
\and
Knut Peterson\textsuperscript{*}%\inst{1}
\and
Quan Tran%\inst{1}
\and
Christian Ricks%\inst{1}
\and
Dhanush Parthasarathy%\inst{1}
\and
Amir Kaidarov%\inst{1}
\and
Neil Deshpande%\inst{1}
\and
Sukaina Najm%\inst{1}
\and
David Han%\inst{1}
\and
Lifeng Zhou\textsuperscript{\dag}%\inst{1}
}
\authorrunning{S. Cai et al.}

\institute{
Drexel University, Philadelphia, USA \\
\email{\{sc3568, kp3275, qdt26, car438, dp3355, amk537, nsd68, sn992, dkh42, lz457\}@drexel.edu}\\
\textsuperscript{*}Equal contribution. \textsuperscript{\dag}Corresponding author.
}

\maketitle

\begin{abstract}
Heterogeneous air-ground robot teams combine complementary sensing modalities, mobility characteristics, and spatial viewpoints that can significantly enhance perception in complex outdoor environments. However, progress in multi-robot collaborative perception has been constrained by the lack of real-world datasets featuring overlapping multi-modal observations from platforms operating in unstructured terrain. We present the \textbf{AGT-CV} (\textbf{A}erial-\textbf{G}round \textbf{T}eam \textbf{C}ross-\textbf{V}iew) dataset, a real-world multi-robot collaborative perception dataset collected using a Clearpath Husky UGV and an Autel EVO~II UAV across diverse unstructured environments, including forest trails, rocky paths, muddy terrain, snow piles, and grass-covered fields. The ground platform provides 3D LiDAR, stereo camera, IMU, and GPS data, while the aerial platform contributes RGB imagery, thermal/infrared observations, and GPS from a complementary overhead viewpoint, allowing for rich cross-modal and cross-view perception. The dataset is collected in 4 unique environments, with over 13,000 synchronized frames across approximately 29 minutes of operation, and includes both SAM~3-based zero-shot segmentation and almost 8,000 manually labeled images. A unique aspect of the dataset is its early-spring collection period, during which sparse tree canopies allow the aerial robot to partially observe the ground robot and terrain through the trees, allowing for occlusion-aware collaborative perception. Unlike prior multi-robot datasets that primarily focus on SLAM or simulated cooperative driving, AGT-CV is specifically designed to support research on cross-view perception, air-ground viewpoint fusion, terrain-aware perception, and collaborative scene understanding in real off-road environments.
\smallskip

\textbf{Dataset:} 
% \href{https://research.coe.drexel.edu/shared/ece/imaple/AGT-CV}{Link}
\href{https://research.coe.drexel.edu/ece/imaple/agt-cv/}{Link}

% (link redacted for double-blind review)
% \href{https://drexel0-my.sharepoint.com/:f:/g/personal/sc3568_drexel_edu/IgAOBo94oaiBRJH3B1bgi4Y_AetpqXDjHcSPw28Rpb-h-lo?e=esDuho}
% {Project Page}

\textbf{Benchmark and Annotator Code:} 
\href{https://github.com/BillCai06/sam3labeler}{Link}
% \href{https://anonymous.4open.science/r/sam3labeler-FD30}{Link}
 
\end{abstract}

\section{Introduction}
\label{sec:intro}

Robot perception in unstructured outdoor environments has advanced rapidly in recent years, driven by progress in deep learning, increasingly capable onboard computing, and rich sensor suites~\cite{wigness2019rugd,jiang2021rellis,mortimer2024goose}. However, most prior perception research has focused on single-robot systems in structured settings such as urban roads and highways. 
% \bl{maybe remove?-In contrast, it is not clear on the contrast part}, 
% In contrast, \textbf{multi-robot collaborative perception}, where heterogeneous robots with different sensing modalities, mobility constraints, and viewpoints jointly observe unstructured environments, remains relatively underexplored. 
In contrast to the availability of single-robot perception datasets, \textbf{multi-robot collaborative perception} remains relatively underexplored. This is especially the case for heterogeneous robot teams with different sensing modalities, mobility constraints, and viewpoints.
Such collaboration can improve robustness, spatial coverage, and sensing redundancy compared with any single-platform solution, making it especially attractive for challenging field environments.

Among heterogeneous robot teams, the combination of Unmanned Ground Vehicles (UGVs) and Unmanned Aerial Vehicles (UAVs) is particularly compelling. Ground robots such as the Clearpath Husky can carry heavier sensing payloads including 3D LiDAR, providing dense geometric observations of the local environment. Small aerial platforms, by contrast, are constrained by size, weight, and power (SWaP) limitations and typically rely on lightweight sensors such as cameras and Inertial Measurement Units (IMUs), yet they offer a broader overhead perspective that is unavailable from the ground. This air-ground complementarity is especially valuable in unstructured outdoor environments with severe occlusions and irregular terrain, where a single viewpoint is often insufficient for reliable perception.

Despite growing interest in multi-robot systems, most existing multi-robot datasets are designed primarily for Simultaneous Localization and Mapping (SLAM)~\cite{chang2022lamp,tranzatto2022team,zhu2023graco,feng2022s3e,tian2023resilient}, emphasizing broad spatial coverage rather than overlapping views that support perception fusion. Many multi-agent perception datasets are also generated in simulation~\cite{sivaprakasam2024tartandrive2,tartanground2025}, which does not capture the sensor noise, calibration drift, and environmental disturbances of real deployments. In autonomous driving, vehicle-to-vehicle (V2V) and vehicle-to-infrastructure (V2X) datasets~\cite{xu2022opv2v,arnold2020cooperative,yu2022dair,li2022v2x} have advanced cooperative perception, but they remain restricted to on-road scenarios. % with homogeneous mobility and Bird's Eye View (BEV) assumptions. 
The CoPeD dataset~\cite{zhou2024coped} represents an important step toward real-world heterogeneous air-ground perception, yet it focuses on relatively structured environments such as indoors or paved outdoor regions, with minimal unstructured dataset content.
%open environments \textcolor{red}{What is open environment? May be better to use terms that contrast our environments.}. In contrast, our dataset targets cluttered off-road settings with dense vegetation, irregular terrain, richer surface variation, thermal sensing, control logs, and substantial manual perception annotation, all of which are important for studying collaborative perception in real field conditions.

Meanwhile, datasets tailored to off-road or unstructured environments have largely centered on single ground robots, without aerial viewpoints or heterogeneous collaboration~\cite{wigness2019rugd,jiang2021rellis,mortimer2024goose,sivaprakasam2024tartandrive2,m2p2_2024}. To the best of our knowledge, no existing dataset simultaneously provides: (i) real-world heterogeneous air-ground multi-robot data collected across diverse off-road environments, (ii) mixed terrain and surface conditions such as mud, grass, shoreline, forest, and snow piles, and (iii) overlapping multi-modal sensor streams including thermal imagery. The thermal modality is particularly valuable in this setting because it can improve robustness under strong illumination changes, help distinguish biological targets such as animals or pedestrians, and reveal water-related or visually ambiguous regions that may be difficult to identify reliably from RGB imagery alone. 

To address this gap, we present the \textbf{AGT-CV} (\textbf{A}erial-\textbf{G}round \textbf{T}eam \textbf{C}ross-\textbf{V}iew) dataset, a heterogeneous multi-robot perception dataset collected in diverse unstructured outdoor environments. Our platform pairs a Clearpath Husky UGV equipped with a stereo camera, 3D LiDAR, IMU, and GPS with an Autel EVO~II UAV carrying RGB, thermal, IMU, and GPS sensors. The ground robot is teleoperated through challenging terrain including muddy ground, grassy fields, gravel paths, and rough hiking trails, while the aerial robot provides a complementary overhead viewpoint for observing scene structure, terrain layout, and robot-environment interactions. An overview of the sensing platforms, collected modalities, and segmentation annotations is shown in Fig.~\ref{fig:teaser}. In addition to synchronized sensor streams, the dataset also records joystick control commands, enabling future study of demonstration-based driving policies and perception-conditioned visuomotor learning in off-road environments. Distinctively, because of the early-spring collection period, the trees present during data collection were largely bare of leaves, allowing the UAV to partially observe the ground scene and UGV through the trees, with occlusions and environmental complexity. As a result, the dataset provides a unique opportunity to study collaborative perception under partial visibility, where aerial sensing can complement ground-level blind spots, and vice versa. A summary of the contributions of this paper are as follows:

\begin{figure*}[t]
    \centering
    \includegraphics[width=1.0\textwidth]{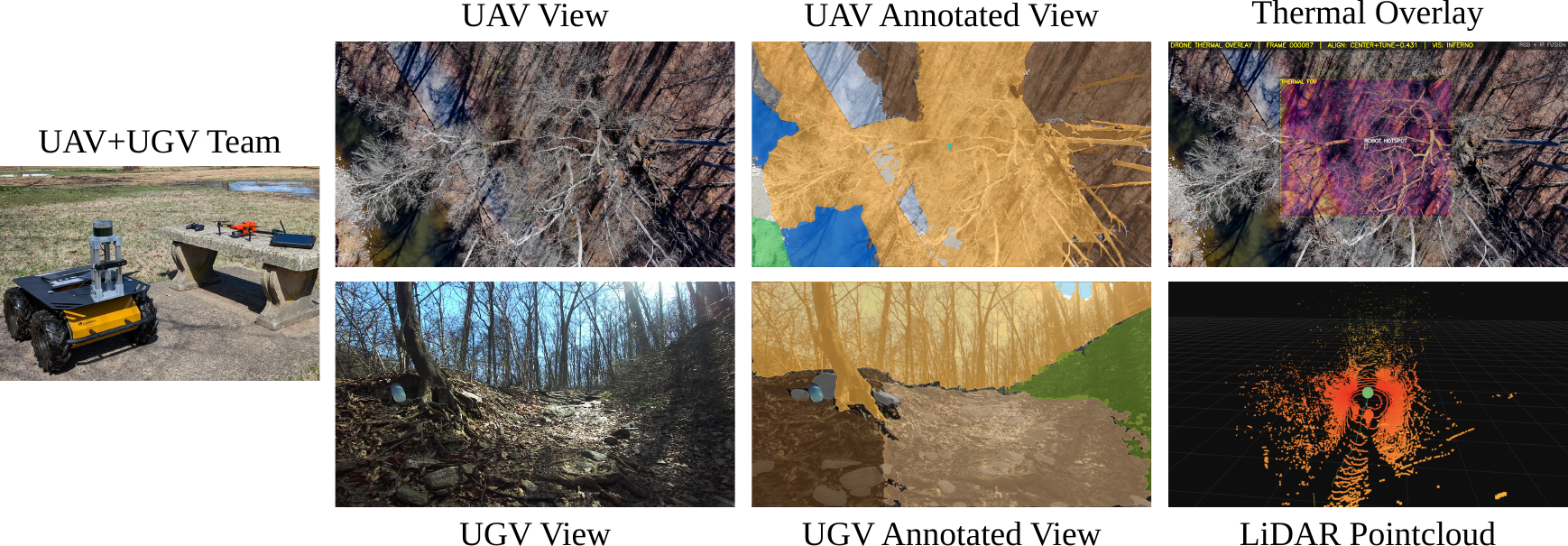}
    \caption{Multi-robot collaborative data collection. The heterogeneous aerial and ground robots collect complementary views, including thermal, RGB, GPS, and LiDAR data. RGB images are annotated with dense class segmentations.}
    \label{fig:teaser}
\end{figure*}

\begin{itemize}
    \item We introduce \textbf{AGT-CV}, a real-world heterogeneous air-ground multi-robot dataset for collaborative perception in diverse off-road unstructured environments. The dataset provides complementary aerial and ground viewpoints with 3D LiDAR, stereo and RGB cameras, thermal imagery, IMU, GPS, and joystick control inputs from both platforms.
    \item We provide accurate pose estimation by fusing KISS-ICP~\cite{vizzo2023kiss} LiDAR odometry with GPS measurements, together with high-quality perception annotations comprising SAM~3-based~\cite{carion2025sam} zero-shot segmentation and 8,000 manually labeled images.
    \item We design the dataset to support concrete downstream tasks enabled by heterogeneous air-ground sensing, including cross-view semantic prediction, path planning with synchronized UAV--UGV context, and broader collaborative scene understanding beyond the BEV assumptions common in existing cooperative perception research.
    \item By recording synchronized teleoperation commands alongside perception data, the dataset also supports emerging directions such as learning from demonstration and visuomotor policy learning for off-road robot navigation.
\end{itemize}

\section{Related Work}

\subsection{Off-Road Robotic Perception Datasets}

Datasets for off-road robotic perception have played an important role in advancing autonomy in unstructured outdoor environments. RUGD~\cite{wigness2019rugd} is an early benchmark that highlights the visual complexity of natural terrain from the viewpoint of a ground robot, with a strong emphasis on semantic segmentation and terrain understanding in unstructured scenes. RELLIS-3D~\cite{jiang2021rellis} further expands this direction by introducing a richer multimodal benchmark with RGB, LiDAR, and semantic annotations for challenging outdoor environments. These datasets have been highly valuable for studying perception and scene understanding in off-road settings, especially from the perspective of a single ground platform. However, they do not address heterogeneous multi-robot perception, aerial-ground viewpoint complementarity, or collaborative sensing across distributed robotic platforms.

\subsection{Heterogeneous and Air-Ground Multi-Robot Datasets}

Several recent datasets have begun to explore heterogeneous robot teams and collaborative sensing. CoPeD~\cite{zhou2024coped} is a notable real-world dataset for collaborative perception with multiple ground robots and an aerial robot, providing an important step toward heterogeneous air-ground perception in realistic settings. The Istanbul Technical University (ITU) heterogeneous robot team dataset~\cite{aybakan20193d} also considers aerial-ground robotic platforms and provides 3D sensing data useful for mapping and localization research. These efforts demonstrate the growing importance of multi-robot sensing beyond single-platform autonomy. However, existing heterogeneous datasets are typically less focused on dense off-road scene understanding in cluttered natural environments, and they provide limited support for studying terrain-centric semantics under vegetation occlusion, mixed surface conditions, and strong viewpoint differences between UAV and UGV observations.

\subsection{Air-Ground Cooperation and Cross-View Robotics}

Another closely related direction studies cooperative air-ground robotics, especially for localization, navigation, and viewpoint assistance. Cristofalo et al.~\cite{cristofalo2016localization} show how UAVs can support UGV localization in GPS-denied environments, illustrating the practical value of complementary aerial and ground viewpoints. More broadly, this line of work motivates the use of UAVs as mobile sensing platforms that can extend perception beyond the limited field of view of a ground robot. However, most prior efforts in this area are primarily algorithmic, focusing on specific localization or navigation methods rather than providing broadly usable benchmark datasets for collaborative perception, cross-view understanding, and multi-modal scene interpretation.

\subsection{Aerial Vision Benchmarks}

Large-scale aerial vision benchmarks such as VisDrone~\cite{zhu2021detection} or the LRDD datasets ~\cite{rouhi2024_LRDD,rouhi2024_LRDDv2,peterson2026_LRDDv3} have significantly advanced UAV-based perception research, especially in object detection and aerial scene analysis. These benchmarks are valuable for developing robust aerial visual models, but they are not designed for joint UAV--UGV robotic perception. In particular, they do not provide synchronized air-ground observations, ground robot trajectories, or terrain-centric multi-modal sensing for collaborative robotic autonomy in unstructured environments.

\subsection{Position of This Work}

Our work lies at the intersection of off-road robotic perception, heterogeneous multi-robot sensing, and air-ground collaborative robotics. Compared with prior off-road datasets, we incorporate complementary aerial and ground viewpoints rather than relying on a ground platform alone. Compared with existing heterogeneous multi-robot datasets, we place greater emphasis on cluttered off-road environments with dense vegetation, shoreline regions, mixed terrain conditions such as mud, grass, water-adjacent areas, and snow piles, as well as partial visibility through sparse early-spring tree canopies. In addition, our dataset includes thermal imagery, synchronized GPS-grounded observations, teleoperation control logs, and 8{,}000 manually labeled images together with zero-shot segmentation annotations. These characteristics make our dataset particularly suitable for studying cross-view semantic understanding, terrain-aware perception, air-ground path planning, perception-conditioned control, and broader collaborative autonomy in complex outdoor environments.

\section{Dataset}
\subsection{Robots and Equipment}

\begin{figure*}[t]
    \centering
    \includegraphics[width=1.0\textwidth]{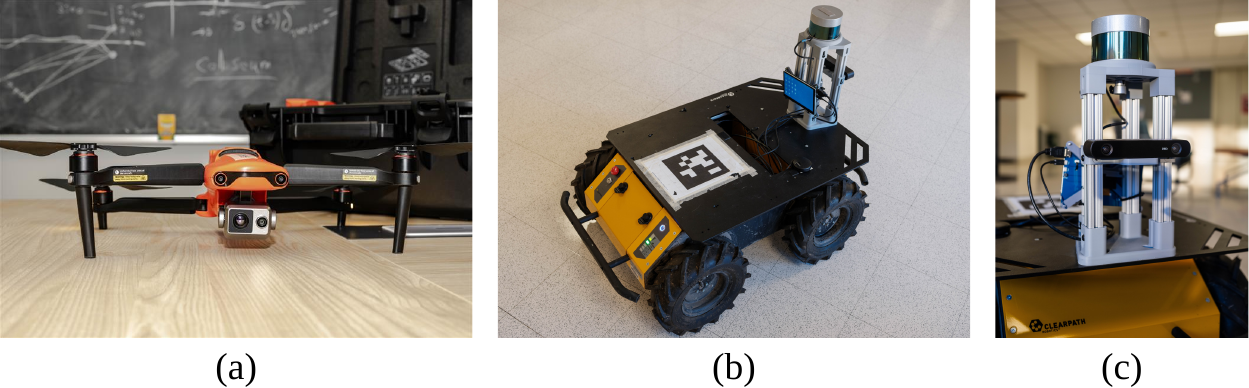}
    \caption{The equipment used for data collection. (a) Autel Robotics EVO II Dual 640T V3, (b) Clearpath Husky A200, (c) Husky sensor tower, including a Velodyne VLP-32C Puck 32 LiDAR, a ZED 2 stereo camera with a built-in IMU. }
    \label{fig:robots}
\end{figure*}
% \vspace{-2mm}
Our dataset is collected using one ground robot and one aerial robot operating in real off-road environments, as shown in Fig.~\ref{fig:robots}. The hardware and sensing configurations of both platforms are summarized in Table~\ref{tab:GA3T_sensor_spec}. The ground platform is a Clearpath Husky A200 UGV equipped with a Velodyne VLP-32C Puck 32 LiDAR, a ZED 2 stereo camera with a built-in IMU, and a Garmin GPS receiver. Onboard computation is provided by an NVIDIA Jetson Orin NX 16GB running ROS~1 Noetic. During data collection, the ground robot is teleoperated using a Sony PS5 controller connected via Bluetooth, and the control commands are logged together with the sensor streams.

The aerial platform is an Autel Robotics EVO II Dual 640T V3 thermal drone. It provides both RGB and thermal aerial observations together with GPS measurements. The visible camera uses a high-resolution CMOS sensor and supports video capture up to 8K resolution, while the thermal camera is based on an uncooled VOx microbolometer with a resolution of $640\times512$.
The drone has a maximum rated flight time of 38~min and includes an omnidirectional binocular sensing system for obstacle awareness. As summarized in Table~\ref{tab:GA3T_sensor_spec}, the two platforms provide complementary sensing capabilities: the UGV captures terrain-level geometric and visual observations, while the UAV supplies broader overhead RGB and thermal context.

To align the aerial and ground data streams in both time and space, we use GPS measurements available on both platforms as a shared reference. The Clearpath Husky A200 is equipped with a Garmin GPS receiver, while the Autel EVO II Dual 640T V3 provides GPS telemetry together with the captured aerial imagery. All sensor data are timestamped using GPS time, which improves the consistency of associating asynchronous measurements from heterogeneous sensors such as LiDAR, stereo cameras, IMU, aerial RGB imagery, and thermal imagery. Although a hardware-level trigger is not utilized, the maximum synchronization offset is maintained within 200~ms. Considering the maximum speed of the Husky was 1m/s and the ground speed of the UAV approximately matched the speed of the Huskly to maintain visual tracking, the offset distance due to the synchronization error was quite small. Errors are still present however, so to compensate for this, we provide temporally-aligned data pairs by interpolating the high-frequency UGV odometry and LiDAR frames to match the UAV video timestamps, ensuring multimodal consistency for perception tasks. GPS also provides a common global frame for coarse spatial alignment between UAV and UGV trajectories, enabling aerial observations to be matched with nearby ground-level measurements during post-processing.
Although this approach does not aim for centimeter-level relative localization, it provides a practical solution for organizing multimodal air-ground data in large-scale outdoor environments.

\begin{table*}[t]
    \centering
    \caption{Sensor and computing specifications of the ground and aerial platforms in AGT-CV.}
    \setlength{\tabcolsep}{3.5pt}
    % \small
    \fontsize{7pt}{7pt}\selectfont
    \begin{tabular}{>{\centering\arraybackslash}p{0.11\linewidth}
                    >{\centering\arraybackslash}p{0.15\linewidth}
                    >{\centering\arraybackslash}p{0.17\linewidth}
                    >{\centering\arraybackslash}p{0.29\linewidth}
                    >{\centering\arraybackslash}p{0.18\linewidth}}
        \toprule
        Type & Model & Key Specs & Specifications & Accuracy / Notes \\
        \midrule

        \multicolumn{5}{l}{\textbf{Ground Platform}} \\
        \midrule
        LiDAR & Velodyne VLP-32C &
        32 channels &
        Range up to $200~\si{m}$, $360^\circ$ horizontal FoV, $40^\circ$ vertical FoV, $5$--$20$ Hz &
        Range accuracy up to $\pm 3~\si{cm}$ \\
        \midrule

        Stereo Camera & Stereolabs ZED 2 &
        Stereo baseline $120~\si{mm}$ &
        RGB up to $2208\times1242$, FoV $110^\circ$(H)$\times70^\circ$(V)$\times120^\circ$(D), image rate up to $100$ fps &
        Depth accuracy $<0.8\%$ at $2~\si{m}$, $<4\%$ at $12~\si{m}$ \\
        \midrule

        IMU & Integrated ZED 2 IMU &
        Accelerometer $\pm 8$ g &
        Gyroscope $\pm 1000$ dps, $400$ Hz &
        Gyro sensitivity error $\pm 0.4\%$ \\
        \midrule

        GPS & Garmin GPS 18x 5Hz &
        WAAS-enabled GPS receiver &
        NMEA 0183 output, $5$ Hz &
        Position acc. $<3~\si{m}$ \\
        \midrule

        Computer & NVIDIA Jetson Orin NX 16GB &
        Onboard computer &
        Running ROS~1 Noetic &
        -- \\
        \midrule

        Tele-operation & Sony PS5 Controller &
        Bluetooth controller &
        Joystick commands logged with sensor streams &
        Event-based \\
        \midrule

        \multicolumn{5}{l}{\textbf{Aerial Platform}} \\
        \midrule
        RGB Camera & Autel EVO II Dual 640T V3 visible camera &
        $0.8''$ CMOS, 50 MP &
        Aperture $f/1.9$, \textbf{23 mm eq. lens}, 
        FoV $85^\circ$, up to $8$K video &
        -- \\
        \midrule

        Thermal Camera & FLIR LWIR thermal camera &
        Uncooled VOx microbolometer &
        $12~\si{\micro m}$ pixel pitch, $8$--$14~\si{\micro m}$ wavelength, $640\times512$, FoV $33^\circ$(H)$\times26^\circ$(V), $30$ fps &
        Temperature accuracy $\pm 3^{\circ}\mathrm{C}$ or $\pm 3\%$ \\
        \midrule

        GNSS / Telemetry & Autel onboard GNSS &
        GPS telemetry &
        Used for geo-referencing and synchronization &
        -- \\
        \midrule

        Obstacle Sensing & 
        Binocular sensing system &
        %Autel binocular sensing system &
        Omnidirectional sensing &
        Obstacle awareness sensing &
        -- \\
        \midrule

        Flight Platform & Autel EVO II Dual 640T V3 &
        Maximum rated flight time &
        $38$ min &
        -- \\
        \bottomrule
    \end{tabular}
    \label{tab:GA3T_sensor_spec}
\end{table*}

\subsection{Dataset content}

\begin{table}[t]
\caption{Overview of dataset contents. Data was collected at four locations, resulting in a total of 13,326 Husky and Drone images and metadata, including 8k manually annotated images and a wide range of traversed terrains.}
\centering
\fontsize{8pt}{8pt}\selectfont
\renewcommand{\arraystretch}{1.4}
\label{table:dataset_overview}
\begin{tabular}{>{\centering\arraybackslash}m{0.16\linewidth}
                >{\centering\arraybackslash}m{0.12\linewidth}
                >{\centering\arraybackslash}m{0.12\linewidth}
                >{\centering\arraybackslash}m{0.12\linewidth}
                >{\centering\arraybackslash}m{0.17\linewidth}
                >{\centering\arraybackslash}m{0.24\linewidth}}
\toprule
\textbf{Dataset Folder}  & \textbf{\# Frames} & \textbf{Data Length} & \textbf{Area (m$^2$)} & \textbf{Annotation Type} & \textbf{Terrains Driven On} \\ \hline

Forest Trail (20260318 \_142502) & 2044 & 264.93 sec. & $\approx 3,464$ & SAM3 & Dirt, rocks, mud, underbrush \\
\hline
Forest Road (20260318 \_141807) & 1702 & 217.50 sec. & $\approx 8,308$ & SAM3 & Gravel road, hiking trail, mud        \\
\hline
Riverbank (20260318 \_140904) & 1582 & 206.89 sec. & $\approx 4,513$ & SAM3 & Rock path, gravel road, concrete, mud \\
\hline
Field (20260320 \_125939)  & 7998 & 1,030.87 sec. & $\approx 13,616$ & Manual & Grass, mud, mulch, metal dock, snow  
\end{tabular}
\end{table}

%\textcolor{red}{TODO: clarify why the UAV and UGV collect data simultaneously (cross-view transformation, etc)}
Our outdoor sequences were collected across three types of real-world off-road environments in the northeastern United States using a ground robot and an aerial robot. Unlike structured outdoor scenes, these environments contain uneven terrain, dense vegetation, natural obstacles, and substantial uncertainty in traversability. During data collection, the Husky is driven by a human operator to traverse the environment, while the aerial drone is flown by a human pilot to record broader overhead RGB and thermal scene context. This setup allows the dataset to capture complementary air-ground viewpoints in realistic off-road conditions without assuming autonomous coordination between the two platforms. The simultaneous data collection approach ensures that the ground robot is present in the UAV camera view, simulating live tracking in a team deployment operation. This allows for potential applications such as tracking robot trajectories in the UAV view, or aiding ground view semantic segmentation with cross-view information. An overview of the dataset contents is shown in Table \ref{table:dataset_overview}. Subfolders are named according to the time of recording in \texttt{<YYYY><MM><DD>\_<hh><mm><ss>} format.

The first environment is a narrow forest trail with dense canopy, steep slopes, rocks, and irregular obstacles. The ground robot experiences highly uneven terrain, while the aerial robot captures a broader view but is often occluded by trees, resulting in complementary yet only partially overlapping perspectives. The second environment is a flatter forest path with dense vegetation, ditches, and holes. Although less steep, it presents ambiguous traversability as vegetation obscures some parts of the terrain, and holes in the path are not easily identifiable due to complex lighting and undergrowth. %visually similar regions may correspond to different ground conditions, making perception challenging. %\textcolor{red}{Not sure what this means}
The third environment is a rocky riverbank with large stones, gravel, and dirt. With less tree cover, the robot is more visible from above, but traversability varies due to differing surface traction and obstacles like large rocks. The fourth environment is a muddy field with sparse trees, puddles, gullies, snow patches, and vegetation.
Notably, the muddy areas of the terrain included deep mud trenches from vehicle treads, which were so extreme that the Husky robot nearly flipped over when traversing them. This highlights the key challenges of the terrain, with a wide mix of easy and difficult scenarios. 

Overall, these environments capture key challenges in off-road air-ground perception, including occlusion, partial viewpoint overlap, uneven and deformable terrain, and ambiguity in traversability, supporting research in realistic, human-operated settings.

\subsection{Data Processing and Annotation}
To prepare the dataset for cross-view association, multimodal fusion, and semantic labeling, we perform three stages of processing. First, we refine the UGV trajectory by aligning LiDAR odometry with GPS measurements to obtain a globally consistent ground reference. Second, we align the UAV RGB and thermal image streams to enable multimodal aerial analysis and visualization. Third, we generate semantic segmentation annotations through a foundation-model-assisted human-in-the-loop pipeline.

\subsubsection{UGV Trajectory Refinement and Ground Truth Generation}

To obtain a globally consistent trajectory for the ground platform, we refine the UGV poses by combining LiDAR odometry with GPS measurements. We first estimate local motion using KISS-ICP~\cite{vizzo2023kiss}, which provides low-drift relative odometry from the 3D LiDAR stream. We then align the LiDAR trajectory to the global coordinate system using GPS measurements.

Specifically, the raw GPS coordinates (latitude and longitude) are converted into a local East-North-Up (ENU) metric frame. Let $\mathbf{k}_i$ denote the relative displacement estimated by KISS-ICP and $\mathbf{g}_i$ the corresponding displacement derived from GPS. We solve for the planar rotation $\theta$ that minimizes the residual error:
\begin{equation}
\min_{\theta}\sum_i \left\| R(\theta)\mathbf{k}_i-\mathbf{g}_i \right\|^2,
\end{equation}
where $R(\theta)$ is the $SO(2)$ rotation matrix.

This yields the heading offset between the LiDAR odometry frame and the GPS-referenced global frame, allowing the LiDAR-consistent trajectory to be rotated and anchored to the global coordinate system. The refined poses retain the local consistency of LiDAR odometry while correcting long-term drift with GPS measurements. These geo-referenced poses support downstream tasks such as cross-view association, collaborative mapping, and air-ground spatial matching.

\begin{figure}[t]
    \centering
    \includegraphics[width=1\linewidth]{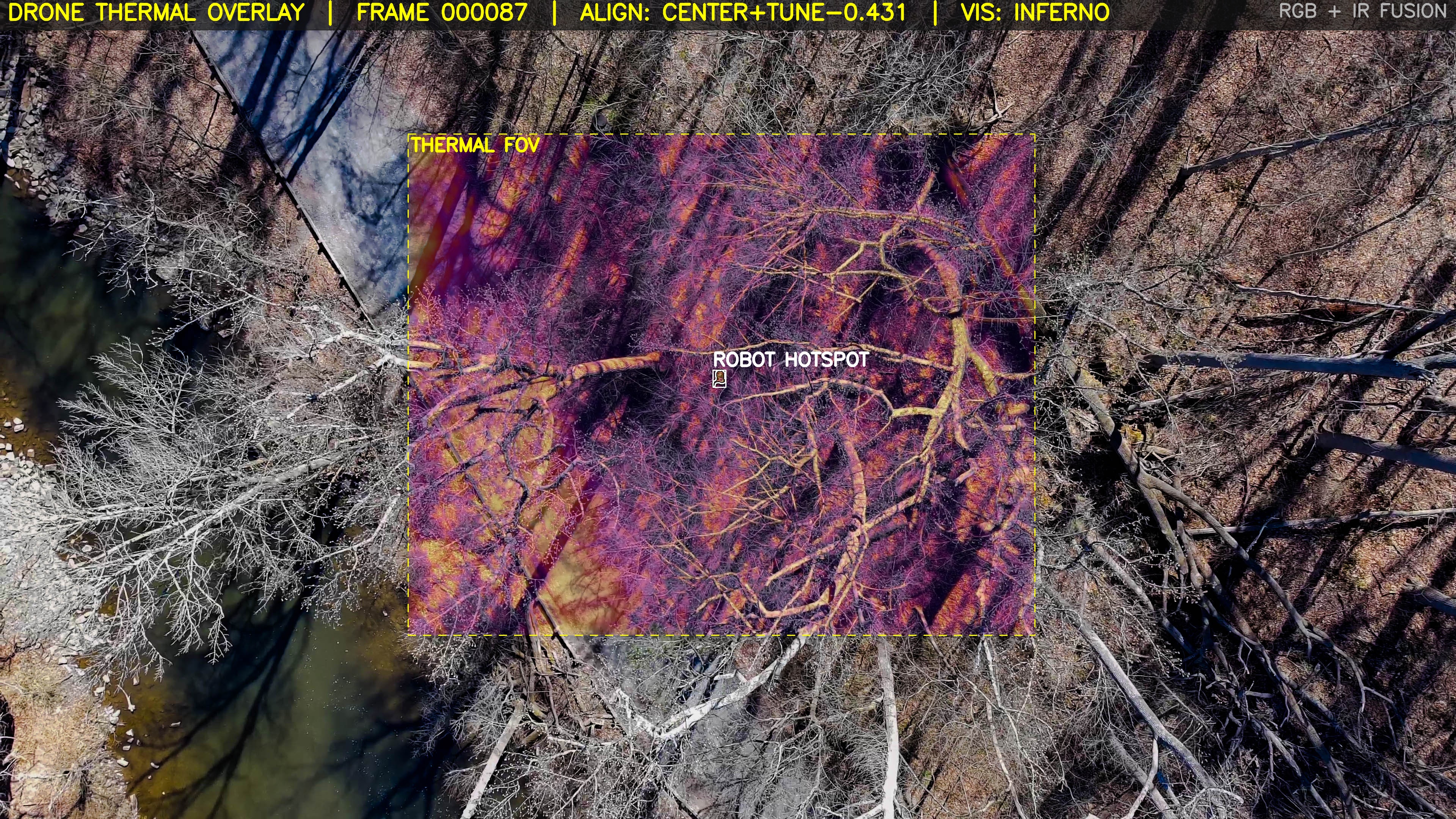}
    \caption{Example of RGB--thermal alignment from the UAV. The thermal image is overlaid on the RGB frame using the \textit{inferno} colormap. The visible hotspot corresponds to the UGV, showing how thermal sensing can help localization under vegetation occlusion.}
    \label{fig:thermal_overlay}
\end{figure}

\subsubsection{Cross-Modal Aerial Alignment}

We align the UAV RGB and thermal streams in the image plane by using the RGB image as the template and warping the thermal image with a gradient-domain matching pipeline. Since no pre-calibrated extrinsic transformation is available, we first apply CLAHE~\cite{pizer1990contrast} and Sobel gradient magnitude extraction~\cite{sobel19683x3} to both modalities to emphasize shared structural boundaries. We then estimate the alignment with a coarse-to-fine search: a coarse pass on downsampled images over scale ratios of approximately $0.3$--$0.8$ of the RGB image width and translations up to $\pm 120$ pixels, followed by full-resolution refinement around the best coarse solution. The alignment is obtained by:
\begin{equation}
\hat{\mathcal{T}} =
\arg\max_{\mathcal{T}}
\left(
\mathrm{NCC}\bigl(\nabla I_{\mathrm{rgb}}, \mathcal{T}(\nabla I_{\mathrm{ir}})\bigr)
- \lambda \, d(\mathcal{T})
\right),
\end{equation}
where $\mathcal{T}$ is a similarity transform with scale and translation, $\mathrm{NCC}(\cdot,\cdot)$ denotes normalized cross-correlation between two gradient images, $d(\mathcal{T})$ is a normalized Euclidean penalty on the offset between the transformed thermal image center and the RGB image center, and $\lambda=0.03$. The aligned streams support thermal overlays such as white-hot and inferno renderings (Fig.~\ref{fig:thermal_overlay}) and help identify the UGV thermal signature in aerial imagery, providing an additional cue for cross-view robot association under vegetation or partial canopy occlusion.

\subsubsection{Foundation-Model-Assisted Segmentation Annotation}

\begin{figure*}[t]
    \centering
    \includegraphics[width=1.0\textwidth]{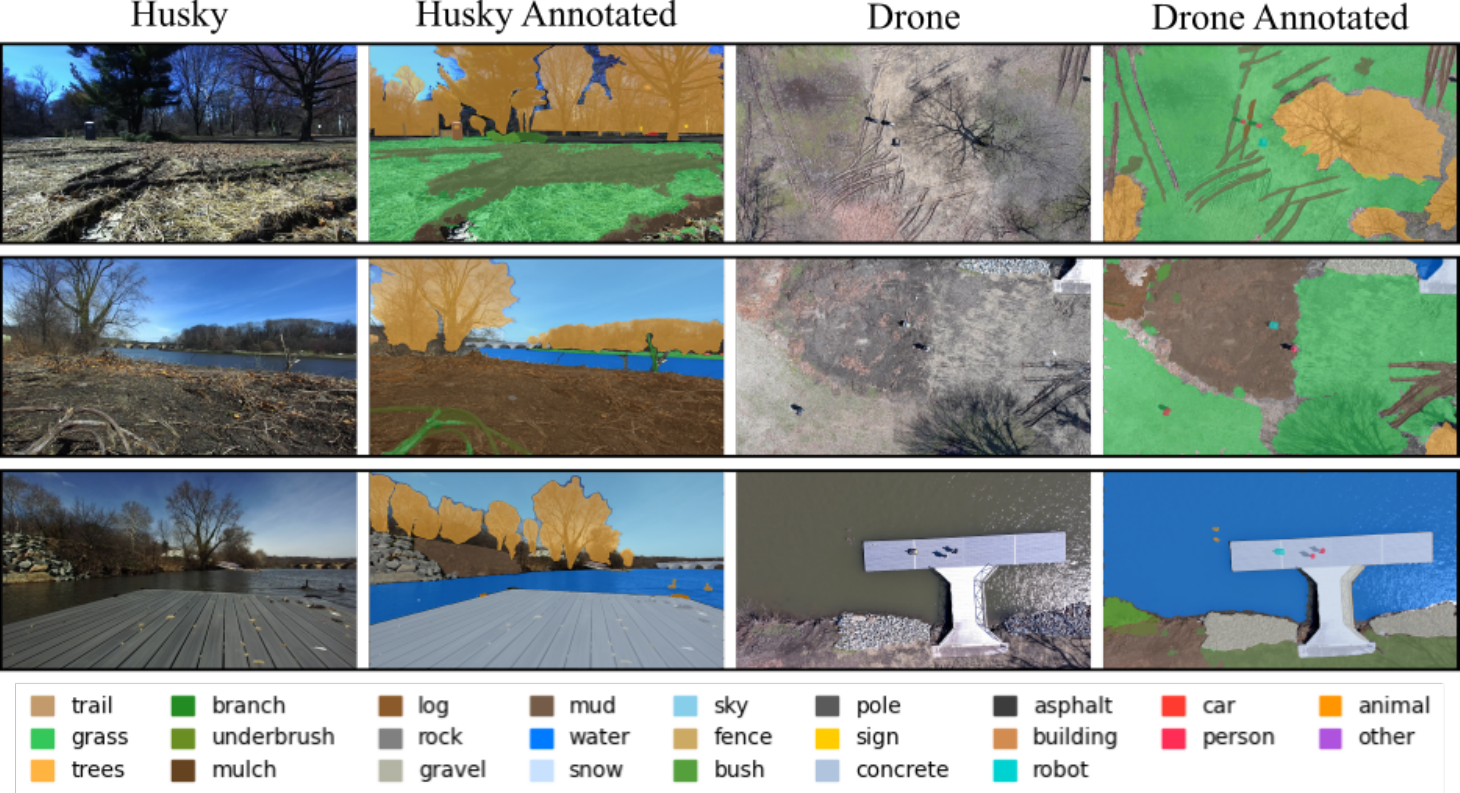}
    \caption{Examples of annotations for paired Husky and Drone images. Images are densely annotated with segmentation labels in the COCO segmentation format. %\textcolor{red}{TODO: tweak figure, make more clear that husky-drone are synchronized (box? replace 2nd one with different view)} \knut{added boxes for now, may adjust more later}
    }
    \label{fig:annotated_example}
\end{figure*}

Our dataset provides semantic segmentation annotations through a foundation-model-assisted human-in-the-loop pipeline. Instead of relying on either fully manual labeling or fully automatic prediction, we first generate mask proposals and then refine them through human correction. In this workflow, annotators primarily assign semantic classes and correct local mask errors, which substantially reduces annotation effort compared with drawing pixel-accurate segmentations from scratch.

We adopt SAM3~\cite{carion2025sam} for mask pre-labeling due to its ability to generate high-quality segmentation masks and support of interactive refinement. Unlike detection-oriented models such as Grounding DINO~\cite{liu2024grounding} which produce bounding-box-level localization, SAM3 is well-suited for pixel-level annotation. This is crucial in off-road environments, where elements like vegetation, mud boundaries, and irregular terrain rarely conform to rectangular priors. By providing accurate initial mask proposals for these complex shapes, SAM3 significantly reduces the manual correction effort required during the annotation process. For label classes we combine the class lists from previous works RUGD \cite{wigness2019rugd} and RELLIS \cite{jiang2021rellis}, and then remove labels that are not present in our data. The full list of labels is shown in Fig.~\ref{fig:annotated_example}.

% \textcolor{red}{TODO: clarify more info on labeling, (define the semantic class taxonomy, explain how ambiguous boundaries between classes, such as mud versus soil, grass versus underbrush, or snow versus sky, were handled, and describe the annotation workflow in greater detail. Information regarding the number of annotators, annotation guidelines, quality-control procedures, and measures of annotation consistency.)}

To support efficient human correction, we built our own annotation tools and service pipeline that allowed annotators to inspect the SAM3-generated masks, assign semantic categories, and correct local errors where needed. This design preserved the efficiency of automatic pre-labeling while maintaining the quality required for the final dataset release.
Using these annotation tools, a team of annotators labeled the images, with one person acting as the lead to clarify ambiguous class taxonomy (e.g. "branch" class is used for distinct large branches, "grass" is used for large fields of a uniform grass plant type, while "underbrush" is used for dense undergrowth of varying plant types, etc.), assist with determining class boundary criteria, and ensure consistency between annotators. All annotations were finally reviewed by the lead to verify their correctness.
The labeling was also completed in phases, where after manual correction of a first set of images, the refined labels were then used to further train an updated SAM3 model adapted to our off-road domain. This iterative process improved the quality of subsequent pre-label generation and further reduced annotation effort in later stages. This approach provided a practical compromise between fully automatic and fully manual annotation, and enabled scalable, high-quality segmentation labeling for challenging scenes. Fig.~\ref{fig:annotated_example} shows examples of annotated UAV-UGV image pairs from the dataset.

\section{Dataset Benchmarking}

To evaluate the benefits of joint aerial--ground training, we fine-tune SAM3 using images from both the UAV and UGV and assess whether the combined data improve terrain segmentation from the challenging aerial viewpoint while preserving performance from the ground viewpoint.

We use the \texttt{Field} subset, which contains approximately 8,000 manually annotated UAV and UGV images. The dataset is divided into training and validation sets using an 85/15 split, with images from both platforms included in each set. Each annotated region is assigned a semantic category from the predefined terrain class list. SAM3 selects from this fixed class list and does not receive free-form text prompts during training or evaluation.

Fine-tuning is performed on $4\times$ NVIDIA H200 GPUs using a two-stage training strategy.

\textbf{Phase 1 (head-only adaptation):}
We freeze the vision encoder and train only the mask decoder for 40 epochs. The learning rate is set to $6\times10^{-4}$ with an effective global batch size of 64. This stage adapts the decoder to the off-road terrain classes and the combined UAV--UGV image domain while preserving the pretrained visual backbone.

\textbf{Phase 2 (backbone fine-tuning):}
Starting from the best Phase 1 checkpoint, we unfreeze the vision encoder and continue training for 20 epochs. The backbone uses a learning-rate multiplier of 0.01, corresponding to $LR_{\mathrm{backbone}}=1\times10^{-6}$. Due to the increased memory requirements, we use a per-GPU batch size of 4 with 8 gradient-accumulation steps, resulting in an effective global batch size of 128.

The model is optimized using the following composite objective:

\begin{align}
L = \lambda_{\text{cls}}L_{\text{focal}} + \lambda_{\text{box}}L_{\text{L1}} + \lambda_{\text{giou}}L_{\text{giou}} 
+ \lambda_{\text{mask}}L_{\text{BCE}} + \lambda_{\text{dice}}L_{\text{dice}} + \lambda_{\text{pres}}L_{\text{pres}}
\end{align}
where $\lambda_{\text{mask}}$ and $\lambda_{\text{dice}}$ are both set to 3.0, and $\lambda_{\text{pres}}$ denotes the presence classification weight.

\subsubsection{Benchmark Results}
We compare the pretrained SAM3 baseline with models jointly fine-tuned on UAV and UGV images from AGT-CV. The results are summarized in Table~\ref{table:benchmarking_results}.

We report class-wise mask IoU by accumulating the intersection and union for each semantic class over the held-out validation set. Let $P_{c,i}$ and $G_{c,i}$ denote the binary predicted and ground-truth regions, respectively, for class $c$ in image $i$. When multiple masks correspond to the same class, they are merged by pixel-wise union before evaluation. The IoU for class $c$ over an evaluation set $\mathcal{D}$ is
\begin{equation}
    \mathrm{IoU}_c =
    \frac{
        \sum_{i \in \mathcal{D}}
        \left|P_{c,i} \cap G_{c,i}\right|
    }{
        \sum_{i \in \mathcal{D}}
        \left|P_{c,i} \cup G_{c,i}\right|
    },
    \label{eq:iou}
\end{equation}
and the mean IoU is computed over the set $\mathcal{C}$ of evaluated semantic classes:
\begin{equation}
    \mathrm{mIoU} =
    \frac{1}{|\mathcal{C}|}
    \sum_{c \in \mathcal{C}} \mathrm{IoU}_c.
    \label{eq:miou}
\end{equation}
We report mIoU separately for the UAV and UGV validation sets to evaluate performance from each viewpoint. We also report a pooled mIoU over the combined UAV and UGV validation images by aggregating the per-class intersections and unions before computing the class-wise IoU.

\begin{table}[t]
\centering
\caption{Class-conditioned mask IoU on the AGT-CV test set for a zero-shot SAM3 baseline and models fine-tuned on AGT-CV using class-prompted segmentation. The overall score is computed by first combining the UAV and UGV results for each semantic class, and then averaging the class-wise IoU values.}

\label{table:benchmarking_results}
\footnotesize
\begin{tabularx}{\linewidth}{@{}lccc@{}}
\toprule
\textbf{Model Strategy} & \textbf{UGV View IoU  } & \textbf{UAV View IoU  } & \textbf{Overall IoU } \\
\midrule
Baseline (Zero-shot)     & 0.7030 & 0.5043 & 0.5615 \\
SAM3 Phase 1 (Head)      & \textbf{0.7429} & 0.6880 & 0.7153 \\
\textbf{SAM3 Phase 2 (Full)} & 0.7314 & \textbf{0.7270} & \textbf{0.7287} \\
\bottomrule
\end{tabularx}
\end{table}
\vspace{-5mm}

\subsubsection{Discussion}
The results show that AGT-CV is a meaningful benchmark for class-conditioned off-road segmentation across heterogeneous viewpoints. The zero-shot SAM3 baseline performs reasonably well on UGV imagery but is noticeably weaker on UAV imagery, reflecting the added difficulty of aerial perception under viewpoint shift, partial canopy occlusion, and weaker local perspective structure. 
% \textcolor{red}{TODO: add a bit more discussion about why UGV IoU decreases during phase 2 training}

Adaptation on AGT-CV improves performance on both views, with the largest gains on the UAV data. This indicates that the dataset captures domain characteristics that are not sufficiently covered by generic pretrained segmentation priors. Head-only adaptation already provides a strong improvement over the zero-shot baseline, while the second stage of backbone fine-tuning further improves UAV and overall performance. These results suggest that AGT-CV supports both effective domain adaptation and challenging evaluation for collaborative air-ground perception in unstructured outdoor environments.

\section{Conclusion}

AGT-CV is a comprehensive collaborative perception dataset tailored for UAV-UGV teams navigating diverse off-road environments. Collected in four unique environments, the dataset contains forest trails, rocky paths, muddy trenches, grass-covered fields, standing water, and large snow piles. With data from a wide number of sensors, including 3D LiDAR, stereo camera, GPS, and RGB+IR imagery, AGT-CV provides a unique set of capabilities with wide range of potential applications. Additionally, the paired ground and aerial camera views include over 8,000 frames with dense semantic segmentation annotations, allowing for terrain identification from multiple viewpoints, with potential applications ranging from multi-view SLAM to terrain-optimized path planning. Together, these features produce a dataset that is uniquely positioned to enable multi-robot collaboration, and stands out as a comprehensive resource for future work in heterogeneous multi-robot coordination.

% commented out for double-blind, will add back in for camera-ready
\section*{Acknowledgments}
Computational resources were provided in part through NSF MRI Award Number 2320600.

\bibliographystyle{splncs04}
\bibliography{ref}
\end{document}